\pdfoutput=1
\documentclass[11pt]{article}

\usepackage[]{acl}

\usepackage{times}
\usepackage{latexsym}

\usepackage[colorinlistoftodos,prependcaption,textsize=tiny]{todonotes}

\usepackage{soul}

\newcommand{\cmt}[1]{}
\usepackage{algpseudocode}
\usepackage{algorithm}
\usepackage{amsmath}
\usepackage{cleveref}
\usepackage{amsfonts}
\usepackage{amssymb} % \varnothing
\usepackage{fdsymbol}
\usepackage{color, colortbl}

\newcommand{\change}[1]{{\textcolor{black}{#1}}}

\definecolor{mypink}{rgb}{0.858, 0.188, 0.478}

\usepackage{booktabs}
\usepackage{multirow}% http://ctan.org/pkg/multirow
\usepackage{hhline}% http://ctan.org/pkg/hhline
\usepackage[T1]{fontenc}
\usepackage[utf8]{inputenc}

\usepackage{microtype}

\title{CoHS-CQG: Context and History Selection for \\ Conversational Question Generation}

\author{
Xuan Long Do$^{1,2,}$\thanks{\quad Contribution during the internship at Institute for Infocomm Research.}, \  Bowei Zou$^{2}$, \ Liangming Pan$^{3}$,  \ Nancy F. Chen$^{2}$, \\ \textbf{Shafiq Joty}$^{1,4}$, \ \textbf{Ai Ti Aw}$^{2}$\\
$^1$Nanyang Technological University, Singapore \\ $^2$Institute for Infocomm Research, A*STAR, Singapore \\
$^3$National University of Singapore, $^4$Salesforce AI Research \\ 
\texttt{\{xuanlong001@e.ntu, srjoty@ntu\}.edu.sg} \\ \texttt{\{zou\_bowei, nfychen, aaiti\}@i2r.a-star.edu.sg} \\ \texttt{liangmingpan@u.nus.edu}
}

\begin{document}
\maketitle
\hfill

\begin{abstract}
Conversational question generation (CQG) serves as a vital task for machines to assist humans, such as interactive reading comprehension, through conversations. 
Compared to traditional single-turn question generation (SQG), CQG is more challenging in the sense that the generated question is required not only to be meaningful, but also to align with the \change{occurred conversation history}. % provided conversation
\change{While} previous studies mainly focus on how to model the flow and alignment of the conversation, there has been no thorough study to date on which parts of the context and history are necessary for the model. 
% Previous studies mainly focus on how to model the flow and alignment of the conversation, but there has been no thorough study to date on which parts of the context and history are necessary for the model. 
We \change{argue} that shortening the context and history is crucial as it can help the model to optimise more on the conversational alignment property. % postulate
To this end, we propose \emph{CoHS-CQG}, a two-stage CQG framework, which adopts a \emph{CoHS} module to shorten the context and history of the input. In particular, \emph{CoHS} selects contiguous sentences and history turns according to their relevance scores by a \emph{top-p} strategy. Our model achieves state-of-the-art performances on CoQA in both the answer-aware and answer-unaware settings. \change{Our work will be publicly available at \url{https://github.com/dxlong2000/CoHS-CQG}.} 

% \bwc{You may release your codes with a git link here.}
\end{abstract}

\section{Introduction}

\begin{figure}[t!]
\vspace{7mm}
\centering
\includegraphics[width=7cm]{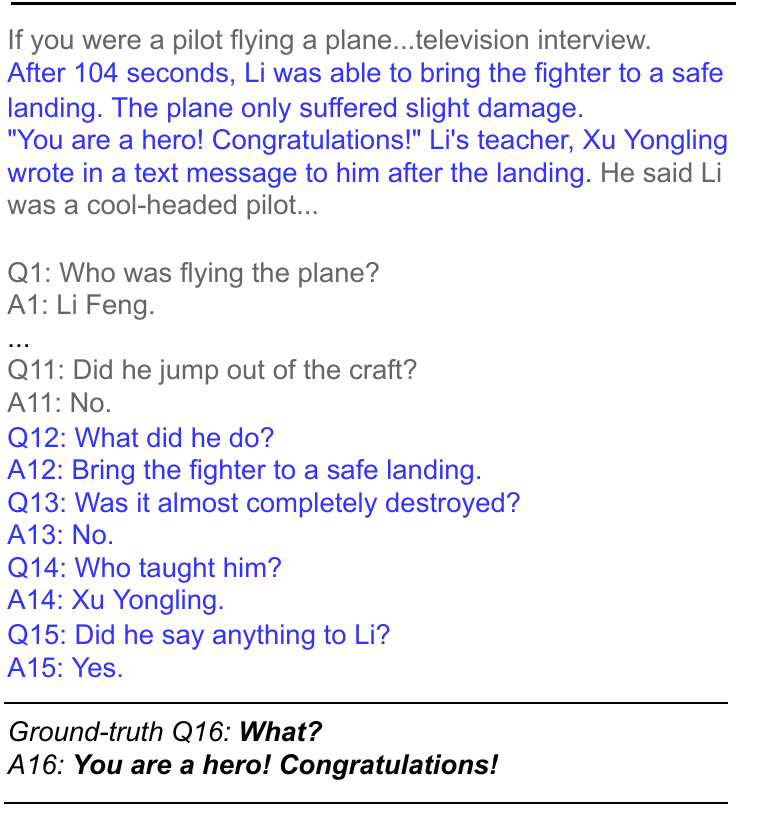}
\vspace{1mm}
{\begin{tabular}{p{7cm}}
\end{tabular}}
\vspace{-5mm}
\caption{\small A dialog sample in CoQA validation set, which reflects not all the sentences in the context and history turns are necessary for \change{generating Q16}. % the $16$-th question generation turn.
}
\vspace{-5mm}
\label{example-real-dialog}
\end{figure}

One of the key goals of AI is to build systems that can understand and assist humans through conversation. In conversations, asking questions is an important dialogue act that serves as an important communication skill for AI models to better interact with humans \cite{James07}. \change{Taking it a step further, asking good questions could facilitate collecting users' intentions and feedback, starting a new topic, and enhancing} the interactivity and persistence of dialogues. In NLP, this line of research is formulated as the task of \textit{Conversational Question Generation (CQG)}, which aims to generate questions based on the conversation history \cite{pan-etal-2019-reinforced,nakanishi-etal-2019-towards}.

Although question generation has been explored intensively \cite{https://doi.org/10.48550/arxiv.1905.08949, lu-lu-2021-survey},
%\bwc{It's better to add another 2\~3 reference here, which are published in 202x.}
most existing studies focus on single-turn question generation, which aims to generate one question from a given context. However, in \change{the scene of} conversation, \change{it} poses \change{an} additional challenge of multi-turn question generation, in which the model is required to generate multiple questions during the conversation, and the generated questions should be coherent and form a smooth conversation flow.

Despite the intensive exploration of single-turn QG, less attention has been drawn on CQG. Previous work of CQG mostly focuses on solving two main challenges: coreference alignment and conversation flow. \citet{gao-etal-2019-interconnected} proposed CFNet to model coreference alignment and conversation flow explicitly. \citet{gu-etal-2021-chaincqg} proposed ChainCQG, a two-stage model with two modules: the Answer Encoder learns the representation of the context and answer in each turn, and the Question Generation learns the representation of the conversational history and \change{generates} \change{the} next turn's question. However, most previous work \change{makes} use of all the context and conversation history \change{indiscriminately}. \change{On the contrary, we argue that} not all sentences in the context, and not all previous turns in the conversation history \change{are} necessary for the model to generate the question in the next turn, \change{and} they may even harm the generation capacity of the model. \Cref{example-real-dialog} shows such an example, where we see that only the blue parts of the context and history are necessary for \change{generating the $16$-th turn's question}. 

\begin{figure*}[t!]
    %  \centering
    % \resizebox{.9\linewidth}{!}{
        % \vspace{mm}
        \includegraphics[height=2.1in,trim={0cm -1cm 0cm 0cm},clip]{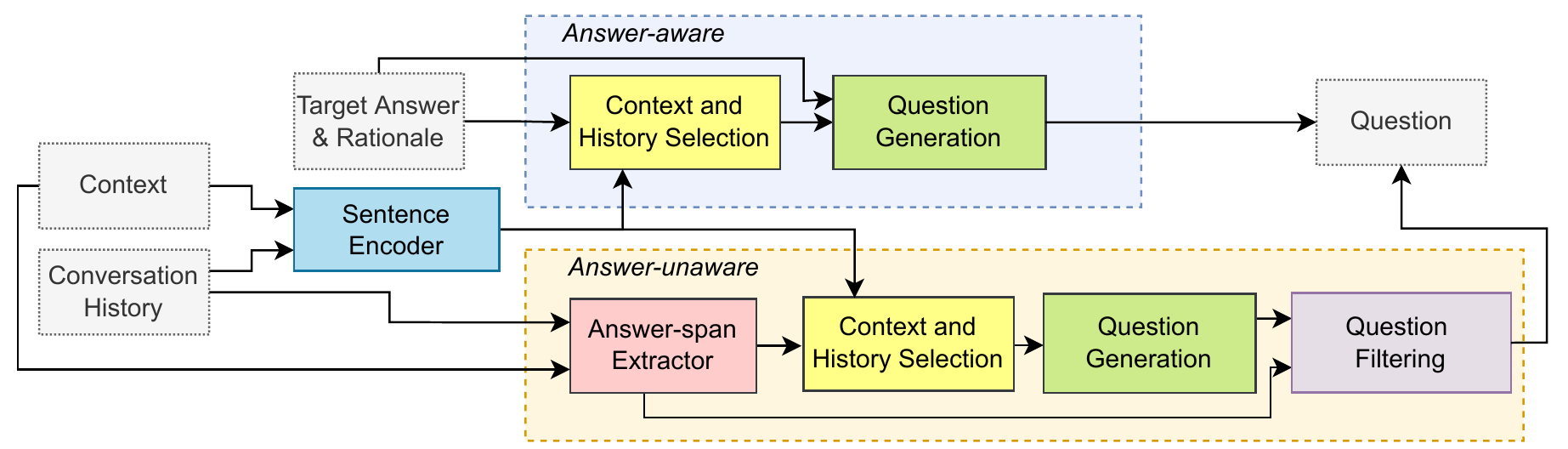}
        \vspace{-7mm}
         \caption{\small \change{An overview of our proposed framework} \emph{CoHS-CQG}. The modules with the same color have the same functionality.
        }
    \label{CoHS-CQG-framework}
    \vspace{-3mm}
    % }
\end{figure*} 
% }

To address the above concerns, we introduce \emph{CoHS-CQG}, a two-stage CQG model, as described in \Cref{CoHS-CQG-framework}. In the answer-aware stage, we input sentences in the context and the conversation history turns into a pretrained sentence-transformer \cite{reimers-2019-sentence-bert} for calculateing the relevance scores of the \change{\texttt{(sentence, history turn)}}
%\bwc{Curly braces usually denote sets. Is it changed to angle brackets or parentheses here?} 
pairs. Context and History Selection (\emph{CoHS}) module (\Cref{answer-aware-framework}) is then employed to shorten the context and conversation history concurrently by selecting $\emph{top-p}$  \{sentence, history turn\} pairs of contiguous sentences in the context, and contiguous previous turns in the history. The shortened context and history are then fed into a T5-based \cite{JMLR:v21:20-074} question generation model to generate the questions. By training the model on the \change{shortened} context and history, we observe that generated questions are generally more aligned with the conversation, which reflects that the model is optimised better in the conversational alignment. Our model achieves state-of-the-art results in the CQG answer-aware setting on both automatic evaluation metrics and a careful human evaluation. In the answer-unaware stage, we propose a pipeline approach \change{(\Cref{answer-unaware-framework})} 
%\bwc{link to Section 3.2} 
to leverage our model on \change{the} answer-unaware \change{setting}, which also \change{achieves} the state-of-the-art \change{performance} on human evaluation. 

In summary, our main contributions are: \textit{(1)} \emph{CoHS-CQG}, a two-stage CQG \change{framework} for \change{both} answer-aware and answer-unaware settings, which adopts a novel module, \emph{CoHS}, to shorten the context and history before inputting them to the QG model. \emph{CoHS} can be plugged into any \change{CQG} model, which makes it easily reproducible, \textit{(2)} new strong state-of-the-art performances on answer-aware and answer-unaware CQG, and \textit{(3)} a thorough analysis and evaluation about the selection capacity of \emph{CoHS}. % and corresponding performance of our model.

\section{Related Work}

\subsection{Single-turn Question Generation}
Single-turn Question Generation (SQG) has been focused extensively through the years. Early studies \change{relied} on syntactic transformation to convert declarative sentences to questions \cite{heilman-smith-2010-good, khullar-etal-2018-automatic}. Recently, \citet{du-etal-2017-learning} showed the limitations of such rule-based methods and formulated the question generation problem as a sequence-to-sequence task. The task is \change{generally} cast into two main streams: answer-aware and answer-unaware.

In the answer-aware setting, the target answer is revealed to \change{SQG models}. The \change{models} then \change{have} to solve the task by either treating the answer as an extra input feature or encoding the answer by a separate network \cite{https://doi.org/10.48550/arxiv.1905.08949}. However, the answer is not available in the answer-unaware case. Traditional approaches in this setting include two main steps: \change{answer-span selection} and answer-aware question generation \cite{du-cardie-2017-identifying, subramanian-etal-2018-neural}. Recent state-of-the-art systems in answer-aware setting \cite{NEURIPS2019_c20bb2d9, qi-etal-2020-prophetnet, 10.1145/3442381.3449892, https://doi.org/10.48550/arxiv.2110.08175} and in answer-unaware setting \cite{scialom-etal-2019-self, https://doi.org/10.48550/arxiv.2005.01107} all rely on transformer-based architectures, and they are commonly evaluated on SQuAD \cite{rajpurkar-etal-2016-squad}. 

\subsection{Conversational Question Generation}
Despite the intensive exploration \change{in both settings of the single-turn QG task}, there is much less exploration in \change{Conversational} Question Generation (CQG). Most of the previous studies focus on \change{the} answer-unaware \change{setting} \cite{pan-etal-2019-reinforced, nakanishi-etal-2019-towards, qi-etal-2020-stay}, but a limited number of works are \change{in} the answer-aware \change{setting}. In general, there are two main challenges in CQG: coreference alignment and conversation flow. Models in the answer-aware setting then have been proposed to solve those problems such as CFNet \cite{gao-etal-2019-interconnected}, by which the coreference alignment and conversation flow are modeled explicitly, and ChainCQG \cite{gu-etal-2021-chaincqg}, which contains two modules: the Answer Encoder (AE) module learns the representation of the context and answer \change{span} in each turn, and the Question Generation (QG) module learns the representation of the conversational history and \change{generates} the next turn's question.

\section{CoHS-CQG} \label{conqg-framework}

We formulate the conversational question generation (CQG) task in two different settings: \emph{answer-aware} and \emph{answer-unaware}. 
For the \emph{answer-aware} CQG, given the referential context $C = \{c_1, c_2, ..., c_m\}$ where $c_i$ is the $i$-th sentence in context, the conversation history \change{$H_n=\{(q_1, a_1), (q_2, a_2), ..., (q_{n-1}, a_{n-1})\}$}, where $(q_i, a_i)$ is the $i$-th turn question-answer pair in conversation, the target answer $a_n$, and the rationale $r_n$, as input $\mathcal{D}^a_n = \{C, H_n, a_n, r_n\}$, the model then learns to generate the question $q_n$. The rationale $r_n$ is an associated text span from the context \change{which contains or explains} the given answer $a_n$.
For the \emph{answer-unaware} CQG, however, given $\mathcal{D}^u_n = \{C, H_n\}$, the model learns to generate the current question $q_n$ without $a_n$ and $r_n$.

\change{Our proposed CoHS-CQG framework is shown in \Cref{CoHS-CQG-framework}. The context $C$ and conversation history $H_n$ are first fed into a Sentence Encoder (SE) to compute the relevance scores. In the answer-aware setting, the relevance scores, together with $a_n$ and $r_n$ are input to the Context and History Selection (CoHS) for selecting the parts of $C$ and $H_n$ that are most relevant to the current generation turn, and they are then input to the Question Generation (QG) \change{module}. In the answer-unaware case, since $a_n$ is unavailable, $C$ and $H_n$ are first fed into the Answer-span Extractor (AE) to extract $a_n$, and $a_n$ is later verified by the Question Filtering (QF) \change{module}.} % We describe the details of these modules below.

\subsection{Answer-aware CQG} \label{answer-aware-framework}

\noindent\textbf{Sentence Encoder (SE)}
Given the context $C$ and conversation history $H_n$, we employ a pretrained sentence-transformer \cite{reimers-2019-sentence-bert} to embed each sentence $c_i$ in $C$, and each question-answer pair $(q_j, a_j)$ in $H_n$ (i.e. the concatenation of $q_j$ and $a_j$), respectively. 
We then compute a relevance matrix $\mathbf{T} \in \mathbb{R}^{|C|\times|H_n|}$ as
\begin{equation}\label{eq:cosine-similarity} 
\mathbf{T}[i][j] = rel(c_i, (q_j, a_j)) = \frac{\mathbf{a}_i\mathbf{b}_j}{|\mathbf{a}_i||\mathbf{b}_j|},
\end{equation}
where $\mathbf{a}_i$ and $\mathbf{b}_j$ are the embeddings of $c_i$ and $concat(q_j, a_j)$, respectively, the relevance score $rel(\cdot)$ is defined as the $cosine$ similarity, and $1 \leq i \leq m, 1 \leq j \leq n-1$.
% \xlong{We then compute a} relevance score of each sentence in the context w.r.t each QA-pair in history, \xlong{which serves as the base for the \emph{CoHS} module to select the most relevant parts of $C$ and $H_n$ w.r.t the current turn.}  \xlong{Specifically, suppose $\mathbf{a_i}$ and $\mathbf{b_j}$ are the embedding vectors of $c_i$ and $concat(a_j, q_j)$ respectively, the relevance score is defined as the the $cosine$ similarity of $\mathbf{a_i}$ and $\mathbf{b_j}$}:
% \begin{equation}\label{eq:cosine-similarity} relevance(c_i, (a_j, q_j)) = \cos{\theta} = \frac{\mathbf{a_ib_j}}{|\mathbf{a_i}||\mathbf{b_j}|} \end{equation}
% Given the context $C$ ($m$ sentences) and the conversation history $H_n$, by computing relevance scores, we obtain the relevance matrix $T_{m \times (n-1)}$ where \begin{equation}\label{eq:similarity-score} T[i][j] = relevance(c_i, (a_j, q_j)) \end{equation}  ($1 \leq i \leq m, 1 \leq j \leq n-1$).

\paragraph{Context and History Selection (CoHS)} \label{CoHS} 
%\paragraph{$\bullet$ Context and History Selection (CoHS)}\label{CoHS} 
% We argue that not all the sentences in the context and the conversation history are necessary for the model to generate the current question. On the contrary, introducing irrelevant noisy may confuse the model. To select the most relevant sentences in $C$ and the most relevant previous utterances in $H_n$, we propose ... 
To generate the current question $q_n$, existing CQG models \cite{gao-etal-2019-interconnected, gu-etal-2021-chaincqg} commonly take the full context $C$ and all the previous question-answer pairs $H_n$ as input. Moreover, in leveraging conversation history, some studies~\cite{ohsugi-etal-2019-simple, zhao-etal-2021-ror-read} have begun to consider how to select historical information related to the current utterance, but only simply selected the last $k$ turns. 
We argue that not all parts of the context and conversation history are necessary for the model to generate the current question since the topic in a conversation may shift. On the contrary, introducing irrelevant parts worsens the performance of the model (See \Cref{tab:main-experiment} and \ref{tab:ablation-previous-turn}). %\bwc{link to the result Table of like ``effects of previous history/context''} \llong{For history I think \Cref{tab:ablation-previous-turn} is the explanation for this statement. However, we haven't made any formal table to report the results if we heuristically to select 3,4,5 sentences in the context. Another option is \Cref{tab:main-experiment} which indicates the effectiveness of our modules I think.}.
% However, it is not always optimum to do so because in different situations, the number of necessary previous turns may vary according to topic shifting, and so are the relevant sentences in context.
%\llong{I think we need to show examples here?} \bwc{yes, could put an example in the ``case study'' section to analyze the problem. just refer it here.}
To address this problem, we propose a \emph{top-p} CoHS strategy that dynamically selects the most relevant sentences in the context concurrently with the most relevant preceding conversation utterances. 
%To address this problem, we propose \emph{top-p} Context and History Selection, a novel algorithm aims to select the most relevant sentences in the context $C$ concurrently with the number of preceding turns dynamically. The algorithm has been done through two modules: \emph{Sentence Encoder (SentE)} and \emph{Context and History Selection (COHS)}. 
% \xlong{We believe that not all the sentences in the context and turns in the conversation history are necessary for the model to generate the next question. Even, inputting the whole context and history may harm the performance, as irrelevant information may confuse the model, as we experiment in \Cref{tab:main-experiment}. Thus, \emph{COntext and History Selection (CoHS)} module aims to select the most relevant sentences in context $C$ and most relevant previous turns in conversation history $H_n$}. 

% Specifically, 
Given the input $\mathcal{D}^a_n=\{C,H_n,a_n,r_n\}$ and the relevance matrix $\mathbf{T}$, CoHS aims to select the \emph{top-p} of sentences and QA pairs from $C$ and $H_n$, respectively. 
Inspired by~\citet{Holtzman2020The}, we formulate our \emph{top-p} CoHS strategy as, finding the subset $C_{sub} = \{c_{v-u}, c_{v-u+1}, ..., c_{v-1}\}$ and $H_{sub} = \{(q_{n-k},a_{n-k}), (q_{n-k+1},a_{n-k+1}), ..., (q_{n-1},a_{n-1})\}$, to satisfy
    \begin{align}
         \label{eq:p1-c1} & minimize(u + k) \\
         \label{eq:p1-c2} & \sum_{i = v-u}^{v-1}\sum_{j = n-k}^{n-1} \mathbf{T}[i][j] \geq p \\
         \label{eq:p1-c3} & (q_{n-1},a_{n-1}) \in H_{sub}, c_s \in C_{sub} 
    \end{align}
    where p is a given threshold, and $c_s$ is the sentence that contains $r_n$.
%\end{problem}
First, the optimizing goal is to minimize the sum of $u+k$, where $u$ and $k$ are the numbers of the contiguous sentences from $C$ and contiguous preceding conversation turns from $H_n$, respectively (Eq.(\ref{eq:p1-c1})). 
Then, the sentences and conversation turns with higher similarity than the threshold $p$ are selected as the candidates for building $C_{sub}$ and $H_{sub}$ (Eq.(\ref{eq:p1-c2})). In addition, since the sentence containing the ground-truth rationale $c_s$ and the last previous conversation turn $(a_{n-1},q_{n-1})$ are intuitively relevant for generating the current question, we set two constraints in Eq.(\ref{eq:p1-c3}). Note that the contiguity of $C_{sub}$ and $H_{sub}$ is necessary due to the integrity and coherence of input. The advantage of the heuristic \emph{top-p} CoHS strategy is that CQG models \change{can} dynamically select the most relevant $C_{sub}$ and $H_{sub}$ according to different conversation progress, which well adapts when topic shifting. When $H_n = \varnothing$, we select five sentences around $c_s$ (see \Cref{appendix:select-five-sentences}).

\paragraph{Question Generation (QG)}
%\paragraph{$\bullet$ Question Generation Model} 
We employ \change{a} T5 \cite{JMLR:v21:20-074} as our question generation model.
To fine-turn \change{the} T5 on the shortened context and history, we concatenate the input $\mathcal{D}^a_n = \{C, H_n, a_n, r_n\}$ in format: \texttt{Answer}: \textit{$a_n$}, \textit{$r_n$} \texttt{Context:} \textit{$C_{sub}$} \texttt{[SEP]} $H_{sub}$. The model then learns to generate the target question $q_n$. 

\subsection{Answer-unaware CQG} \label{answer-unaware-framework}
In \Cref{answer-aware-framework}, we utilize 1) the ground-truth previous conversation history $H_n$, and 2) the ground-truth current answer $a_n$ and rationale $r_n$, to verify how well the model \change{performs in} generating the current question $q_n$. However, in a more realistic scenario such as a dialogue system, it is necessary to verify whether the model has a good ability to generate questions continuously, that is, the coherence and fluency of the generated questions. To this end, we propose an \emph{answer-unaware} process as shown in \Cref{CoHS-CQG-framework}, including \emph{Answer-span Extractor}, \emph{CoHS} (depicted in \Cref{answer-aware-framework}), \emph{QG} (depicted in \Cref{answer-aware-framework}), and \emph{Question Filtering}.

\paragraph{Answer-span Extractor (AE)}
% \paragraph{$\bullet$ Answer-span Extractor} 
First, we treat the earliest sentence in the context as the current rationale $r_n$ such that $r_n$ does not contain any rationales of previous turns. Then, a T5 model is trained on SQuAD \cite{rajpurkar-etal-2016-squad} to predict the target answer span ($a$) given its original sentence in context ($r$). We use the model to extract $a_n$ from $r_n$. Note that we remove the answer spans that are the same as those of previous turns, to ensure that the generated questions are informative enough. 
Finally, we obtain a set of selected candidate answer spans $A_n = \{a^*_1, a^*_2, ..., a^*_t\}$. Each $a^*_i \in A_n$, together with $r_n$, and $\mathcal{D}^u_n$ are fed into the CoHS and QG modules to generate the candidate question $q^*_i$.

\paragraph{Question Filtering (QF)}
% \paragraph{$\bullet$ Question Filtering}
Under the answer-unaware \change{setting}, since the conversation history is not manually-labeled, we observe that one type of the common errors is that the generated question may not be answerable by the given context, or its answer may not the provided target answer $a^*_i$. To address this issue, we train a T5 model on CoQA \cite{reddy-etal-2019-coqa} to answer the generated question $q^*_i$, and only accept $q^*_i$ if the predicted answer is the same as $a^*_i$.

\section{Experimentation}

\begin{table*}
\centering
\resizebox{\textwidth}{!}{%
\begin{tabular}{lccccccc}
\hline
Model & ROUGE-L & B1 & B2 & B3 & B4  & METEOR & BERTScore\\
\hline
% ReDR & - & - & - & - & - & - & - \\
CFNet & 41.25 & 34.24 & 22.71 & 16.57 &  12.39  & 27.76 & 91.43 \\
ChainCQG$^*$  & 42.22 & 35.54 & 26.03 &  19.63 & 14.54 & 30.97 &  92.54\\
BART$_{base}$ & 44.77 & 35.86 & 26.32 & 19.84 & 15.09  & 31.60 & 92.95 \\
T5$_{base}$  & 45.80 & 39.09  & 29.04 & 22.17 & 17.03  & 34.09 & 93.07 \\
\hline
T5$_{base}$ + \emph{dyn-HS (p = 0.5)}  & 48.64 & 40.83 & 30.74 & 23.64 & 18.18 & 36.49 & 93.43 \\
T5$_{base}$ + \emph{dyn-CS (p = 1)}  & 49.69 & 41.62 & 31.44 & 24.29 & 18.72  & 37.42 & 93.61 \\
%T5$_{base}$ + \emph{QS}  & 48.71 & 41.38 & 31.11 & 23.94 & 18.47  & 36.85 & 93.46 \\
\emph{CoHS-CQG (Ours, p = 5)} & \textbf{49.91} & \textbf{42.10} & \textbf{31.86} & \textbf{24.65} & \textbf{19.11} & \textbf{37.76} & \textbf{93.65}\\
\hline
\end{tabular}
}
\caption{\label{tab:main-experiment}
Automated evaluation results on our test set (i.e. CoQA validation set). \emph{dyn-HS} and \emph{dyn-CS} are \emph{dynamic History Selection} and \emph{dynamic Context Selection} respectively (\Cref{ablation:dynamic-one}). B1 to B4 denotes BLEU 1-4.
}
\end{table*}

\subsection{Experimental Settings} \label{ssec:experiment-setting}

\paragraph{Dataset}
We conduct experiments on CoQA \cite{reddy-etal-2019-coqa}, a large-scale CQA dataset including 8k conversations. Each conversation contains a referential context and multiple question-answer pairs. In total, there are 127k question-answer pairs collected via Amazon Mechanical Turk. The key characteristics of this dataset are its factoid questions (\textit{i.e.} What? Where? When? How long?) and free-form answers. Since the test set of CoQA is unavailable, we randomly sample 10\% of the original training set as our new \emph{validation set}, and keep the original validation set as our \emph{test set} so that future works can be compared with us. 

\paragraph{Baseline Models}
We use \change{a} T5$_{base}$ (220M) as \change{our} \emph{CoHS-CQG}'s backbone. For the answer-aware baselines, we reimplement CFNet \cite{gao-etal-2019-interconnected}, an effective CQG framework. We also finetune \change{a} T5$_{base}$ \cite{JMLR:v21:20-074} and \change{a} BART$_{base}$ \cite{lewis-etal-2020-bart}, the SOTA \change{transformer-based} generation models, on CoQA. For the answer-unaware baseline, we compare with the SOTA framework ReDR \cite{pan-etal-2019-reinforced}. 

\paragraph{Implementation Details}
We initialise \emph{CoHS-CQG} with pretrained \change{checkpoints} from Huggingface \cite{wolf-etal-2020-transformers}. We use AdamW \cite{loshchilov2018decoupled} with the warmup ratio of 0.1 and the initial learning rate of 1e-4. We train the model for 100k iterations with standard window size of 512, and use a Beam search decoding strategy with beam size of 4. 

\paragraph{Evaluation Metrics}
We compute the standard $n$-gram-based similarity metrics, which are commonly used for text generation, including ROUGE-L \cite{lin-2004-rouge}, BLEU (1-4) \cite{10.3115/1073083.1073135}, and METEOR \cite{banerjee-lavie-2005-meteor}. We compute BLEU 1-4 by \emph{corpus\_bleu} function from NLTK library.\footnote{\url{https://www.nltk.org/}} We compute ROUGE-F scores in our evaluations by Python implementation of \emph{rouge-score} library.\footnote{\url{https://github.com/google-research/google-research/tree/master/rouge}} We also calculate BERTScore \cite{Zhang2020BERTScore:}, a similarity score between the generated and \change{ground-truth} texts by using deep contextualized embeddings. 

Human evaluation is also important to the CQG task since the CQG model may generate the question for the following turn in multiple ways, given the target answer. As such, we conduct human evaluation on both the answer-aware setting and answer-unaware setting. 

\begin{table}[t!]
\centering
\resizebox{0.45\textwidth}{!}{%
\begin{tabular}{lccc}
    \hline
    Model & Flu. & C-Align & Ans. \\
    \midrule \multicolumn{4}{c}{\textbf{Answer-aware}} \\
    BART$_{base}$  & 2.38 & 2.23 & 2.34\\
    T5$_{base}$  & 2.72 & 2.44 & 2.54\\
    \emph{CoHS-CQG} (\emph{Ours}) & \textbf{2.74} & \textbf{2.58} & \textbf{2.60} \\
    \hline
    Krippendorff's $\alpha$ & 0.80 & 0.71 & 0.77 \\
    \midrule \multicolumn{4}{c}{\textbf{Answer-unaware}} \\
     ReDR  & 1.06 & 1.06 & 1.05\\
     \emph{CoHS-CQG} (\emph{Ours}) & \textbf{2.70} & \textbf{2.41} & \textbf{2.73} \\
    \hline
     Krippendorff's $\alpha$ & 0.84 & 0.85 & 0.82 \\
    \hline
    \end{tabular}
}

\caption{Human evaluation \change{results} on the validation set of CoQA. \textbf{Top}: answer-aware, on 100 random generated samples; \textbf{Bottom}: answer-unaware, on 20 random conversations. ``Krippendorff's $\alpha$'' shows the inter-annotator agreement. \textit{Flu.}: Fluency, \textit{C-Align}: Conversational Alignment, \textit{Ans.}: Answerability. }

\label{tab:human-evaluation}
 \label{tab:human-evaluation-answer-aware}
\vspace{-3mm}
\end{table}

\begin{table}[t!]
\centering
\resizebox{0.48\textwidth}{!}{%
\begin{tabular}{ccc|cc}
\hline
\emph{p value} & Avg. \#S & Avg. \#P & ROUGE-L & BLEU-4 \\
\hline
% 0 & - & - & - & - & -  \\
1 & 3.18 & 1.89  & 48.96 & 18.19  \\
2 & 3.73 & 2.49 & 49.38  & 18.61  \\
3 & 4.27 & 2.95 & 49.41 & 18.65  \\
%4 & 5.03 & 3.33 & 49.71  & 18.86  \\
5 & 5.29 & 3.67 & \textbf{49.91} & \textbf{19.11}  \\
7 & 6.13 & 4.25 & 49.35  & 18.69  \\
10 & 8.12 & 4.92 & 49.18  & 18.52  \\
\hline
\(\infty\) & 16.09 & 7.97 & 45.80 & 17.03   \\
\hline
\end{tabular}
}
\caption{Average \#Sentences, \#Prev. Turns at different \emph{p} values of preprocessed contexts and histories by \emph{CoHS} of CoQA. \(\emph{p} = \infty\) means selecting the full context and history. \textit{Avg. \#S}: average number of sentences, \textit{Avg. \#P}: average number of previous turns. 
}
\vspace{-3mm}
\label{tab:main-ablation}
\end{table}

\subsection{Automatic Evaluation} \label{main:experiment}

\Cref{tab:main-experiment} shows the automatic evaluation results. We observe that \emph{CoHS-CQG (p = 5)} achieves state-of-the-art performance on all the automatic evaluation metrics. In particular, we derive 3 observations. First, \emph{CoHS-CQG} improves its original baseline T5$_{base}$ significantly. Second, comparing to only dynamically selecting previous turns (T5$_{base}$ + \emph{dyn-HS}) or sentences in the context (T5$_{base}$ + \emph{dyn-CS}), \emph{CoHS-CQG} achieves better performances, which indicates that dynamically selecting both is more effective. Third, with the threshold of relevance $p=5$ (Eq.(\ref{eq:p1-c2})), the \emph{CoHS} module shortens the context to around 5 sentences and the history to 3 previous turns on average (\Cref{tab:main-ablation}), by which it achieves the best performance.

We also compare our \emph{CoHS-CQG} with the current SOTA answer-aware CQG model, ChainCQG \cite{gu-etal-2021-chaincqg} which contains two GPT-2 \cite{radford2019language} blocks. Since the provided codes from the authors are incomplete, and the reported results of ChainQCG in \cite{gu-etal-2021-chaincqg} are on the authors' own test set (they splitted 10\% of the training set to become their own test set), we \change{were not} able to reproduce the results. Thus, we reimplement ChainCQG (denote it as ChainCQG$^*$). We can see that \emph{CoHS-CQG} outperforms ChainCQG$^*$ on all automatic evaluation metrics significantly. 

% \liangming{put more findings here?}

\subsection{Human Evaluation}

\paragraph{Evaluation Setup} 
We further conduct human evaluation to validate the results. In answer-aware case, we randomly select 100 generated questions associated with the context and conversation history. In answer-unaware case, however, since there is no ground-truth history, simply evaluating 100 random generated samples may not be a fair comparison. Thus, we first select 20 random contexts in our test set. For each context, since the number of turns generated by our model and the competing one, ReDR \cite{pan-etal-2019-reinforced}, may not be the same, we heuristically select the first five generated turns from each model's output to compare, resulting in 100 samples in total. We hire three annotators who are English native speakers. Each annotator \change{was instructed to} rate the generated questions on a 1-3 scale (3 for \change{the} best) based on three criteria: \textbf{(1) Fluency} measures \change{not only} the grammatical correctness \change{but also the} meaning, and factual correctness of generated questions, \textbf{(2) Conversational Alignment} measures the alignment of generated questions with the given conversation, \textbf{(3) Answerability} measures whether the generated questions are well answerable or not. We measure the annotators' agreement by Krippendorff’s alpha \cite{krippendorff2011computing}. Our rating system is described in \Cref{ssec:human-rating-system}. 

\paragraph{Observations}
The top of \Cref{tab:human-evaluation} shows the averages of human scores over three annotators in the answer-aware setting. We derive two main observations. First, there is a significant improvement in the \emph{Conversational Alignment} of \emph{CoHS-CQG} compared to T5, which indicates that with the shortened context and history as input, the model learns to focus and align with the \change{given} conversation history much better. Second, compared to T5, there is also a slight increase in the \emph{Answerability}, which further shows that the quality of the generated questions is improved. There is also a minor improvement in the \emph{Fluency}, which is reasonable because T5 commonly generates \change{fluently}, grammatically and meaningfully correct questions. Our annotators have a good overall agreement with an alpha coefficient of 0.76.

The bottom of \Cref{tab:human-evaluation} shows the human evaluation for the answer-unaware setting. First, we observe that ReDR has low \emph{Fluency} score due to most of the generated questions are factually wrong or have no meaning \change{associated} with the given context. It also has low scores on the other two metrics \change{as the generated questions are frequently repetitive}. Second, the generated questions by \emph{CoHS-CQG} are generally high-quality, fluent and answerable as they already passed the \emph{Question Filtering} module. The annotators achieve a good overall inter-agreement with an alpha coefficient of 0.83. 

\subsection{Effects of Context and History Selection} \label{ablation:dynamic-one}
We further conduct the studies \change{about} the performance \change{of T5} when we dynamically select the context sentences or the previous turns but not both of them concurrently. In this section, we formulate these two problems as below.

\begin{table}[!t]
\centering
\resizebox{0.4\textwidth}{!}{%
\begin{tabular}{ccc}
\hline
\emph{\#Pre. turns} & ROUGE-L & BLEU-4 \\
\hline
1 & 48.14 & 17.43 \\
2 & \textbf{48.34} & 17.66 \\
3 & 48.21 & \textbf{17.68} \\
4 & 47.77 & 17.64 \\
5 & 47.15 & 17.59 \\
6 & 46.90 & 17.12 \\
\hline
Full history & 45.33 & 16.73 \\
\hline
\end{tabular}
}
%\vspace{-2mm}
\caption{Performance of T5$_{base}$ with different fixed number of previous turns on our validation set.}
\label{tab:ablation-previous-turn}
%\vspace{-3mm}
\end{table}

\paragraph{Dynamic Context Selection}
%\paragraph{$\bullet$ Dynamic Context Selection}\label{dynamic-context}
In this setting, we follow the previous studies on CQA \cite{ohsugi-etal-2019-simple, zhao-etal-2021-ror-read} to select the last $k$ previous turns. The results are shown in \Cref{tab:ablation-previous-turn}. Since T5$_{base}$ achieves the best performance \change{on} BLEU-4 by using the last 3 previous turns, we adopt this setting in the following independent context selection experiments. Given the context $C$, answer $a_n$, rationale $r_n$, and the last $k$ previous turns $H_{sub} = \{h_{n-k}, h_{n-k+1}, ..., h_{n-1}\}$, $h_i$ = $concat(q_i, a_i)$, we formulate the context selection problem in this section as finding the smallest subset $C_{sub} = \{c_{v-u}, c_{v-u+1}, ..., c_{v-1}\}$, to satisfy:
    \begin{align}
        \label{eq:p2-c1} \sum_{x = v-u}^{v-1}\sum_{y = n-k}^{n-1} T[x][y] \geq p \\
        \label{eq:p2-c2} c_s \in C_{sub} 
    \end{align}
    where $p$ is a given threshold, and $c_s$ is the sentence that contains $r_n$.
We name this model as T5$_{base}$ + \emph{dyn-CS} in \Cref{tab:main-experiment} where \emph{dyn-CS} stands for dynamic Context Selection. 

\Cref{tab:ablation-dynamic-context} shows how different values of threshold $p$ (Eq.(\ref{eq:p2-c1})) affects the selection and the performance of the model. We observe that with a fixed number of previous turns $k = 3$, $p=1$ gives us the best performance on ROUGE-L and BLEU-4. By setting threshold $p=1$, and fixed 3 previous turns, the \change{\emph{CoHS} module} selects around 4 sentences in each context sample on average. The result indicates that selecting more contexts does not lead to better performance, which is consistent with our motivation.

\begin{table}[t!]
\centering
\small
\scalebox{1}{\begin{tabular}{cc|cc}
\hline
\emph{p value} & Avg. \#Sentences  & ROUGE-L & BLEU-4 \\
\hline
0 & 1 & 48.15 & 17.88\\
1 & 4.35 & \textbf{49.69} & \textbf{18.72} \\
2 & 8.28 & 49.47 & 18.43\\
3 & 11.51 & 48.45 & 18.30\\
4 & 13.42  & 48.80 & 18.16\\
\hline
\(\infty\) & 16.09 & 48.31 & 18.01 \\
\hline
\end{tabular}
 }
%\vspace{-2mm}
\caption{\small Average number of selected sentences from context (fixing the last 3 previous turns) of each \emph{p} value (left) and performance of ``T5$_{base}$ + \emph{dyn-CS}'' on our test set (right). \(\emph{p} = \infty\) and \emph{p} = 0 denote utilizing the full context and only the sentence that contains the rationale respectively.}
%\caption{\small Avg. \#Retrieved sentences at each \emph{p} value of preprocessed contexts (by our \emph{Context Selection} module) on the whole CoQA (with fixed 3 previous turns) and performances on our test set. \(\emph{p} = \infty\) means selecting the full context.}
\label{tab:ablation-dynamic-context}
%\vspace{-3mm}
\end{table}

\paragraph{Dynamic History Selection}
%\paragraph{$\bullet$ Dynamic History Selection}
In this setting, we follow most previous works on CQA \cite{ohsugi-etal-2019-simple, zhao-etal-2021-ror-read} and CQG \cite{gao-etal-2019-interconnected, gu-etal-2021-chaincqg} to use the whole context $C$, and then we dynamically select different numbers of previous turns. We formulate the history selection problem as finding the smallest subset $H_{sub} = \{(q_{n-k},a_{n-k}), (q_{n-k+1},a_{n-k+1}), ..., (q_{n-1},a_{n-1})\}$, to satisfy:
    \begin{align}
        \label{eq:p3-c1} \sum_{x = 1}^{m}\sum_{y = n-k}^{n-1} T[x][y]  \geq p \\
        \label{eq:p3-c2} (q_{n-1},a_{n-1}) \in H_{sub} 
    \end{align}
    where $p$ is a given threshold, and $c_s$ is the sentence that contains $r_n$.
We name this experiment as T5$_{base}$ + \emph{dyn-HS} in \Cref{tab:main-experiment} where \emph{dyn-HS} stands for dynamic History Selection. 

\Cref{tab:ablation-dynamic-history} shows how different values of threshold $p$ (Eq.(\ref{eq:p3-c1})) affects the selection and performance of the model. We can observe that with the full context, $p=0.5$ achieves the best performance on both ROUGE-L and BLEU-4. By setting the threshold $p=0.5$, the \change{\emph{CoHS} module} then selects around 3 previous turns on average. This observation is in line with our following results in \Cref{tab:ablation-previous-turn} (see \Cref{sec:fix-different-turns}), by which we conclude that with different values of fixed number of previous turns, $k=2$ and $k=3$ \change{achieve} the best results. 

\begin{table}[t!]
\centering
\small
\scalebox{1}{\begin{tabular}{cc|cc}
\hline
\emph{p value} & Avg. \#History Turns & ROUGE-L & BLEU-4 \\
\hline
0 & 0 & 43.91 & 15.32\\
0.5 & 2.57 &  \textbf{48.64} & \textbf{18.18} \\
1 & 4.11 &  48.27 & 18.07   \\
2 & 6.32 &  47.14 & 17.70\\
3 & 7.45 &  46.74 & 17.32 \\
\hline
\(\infty\) & 7.97 &  45.80 & 17.03 \\
\hline
\end{tabular}
 }
% \vspace{1mm}
\caption{\small Average number of selected history turns from conversation (with the full context) of each \emph{p} value (left) and performance of ``T5$_{base}$ + \emph{dyn-HS}'' on our test set (right). \(\emph{p} = \infty\) and \emph{p} = 0 denote utilizing the full history and none of the history respectively.}
% \caption{\small Average \#Retrieved history turns at each \emph{p} value of preprocessed histories (by our \emph{History Selection} module) on the whole CoQA with full context. \(\emph{p} = \infty\) means selecting the full history (i.e without \emph{History Selection}).}
\label{tab:ablation-dynamic-history}
%\vspace{-3mm}
\end{table}

\subsection{Discussion} \label{ablation-studies}
%\subsection{Ablation Study} \label{ablation-studies}

\paragraph{Effects of Relevance Threshold $p$}
To further understand how the threshold $p$ (Eq.(\ref{eq:p1-c2})) controls the selection of context and conversation history, we conduct experiments with different values of $p$. \Cref{tab:main-ablation} shows the average number of the selected sentences and the selected previous turns, together with the performances of T5$_{base}$ \change{on} ROUGE-L and BLEU-4. First, we can observe that on average, the difference between the \#Sentences and \#Pre. Turns is not large for all values of $p$, which reflects that our \emph{top-p} algorithm does not prioritise selecting long context over short history \change{and} vice-versa. This indicates that the relevance scores assist the algorithm to select the context sentences, together with the history turns in a reasonable way. Second, with $p=5$, T5$_{base}$ yields the best performance, as we discussed in \Cref{main:experiment}. 

\paragraph{Effects of Different Fixed Previous Turns} \label{sec:fix-different-turns}

\begin{figure*}[t!]
    %  \centering
    \includegraphics[height=6.7in,trim={0cm -1cm 0cm 0cm},clip]{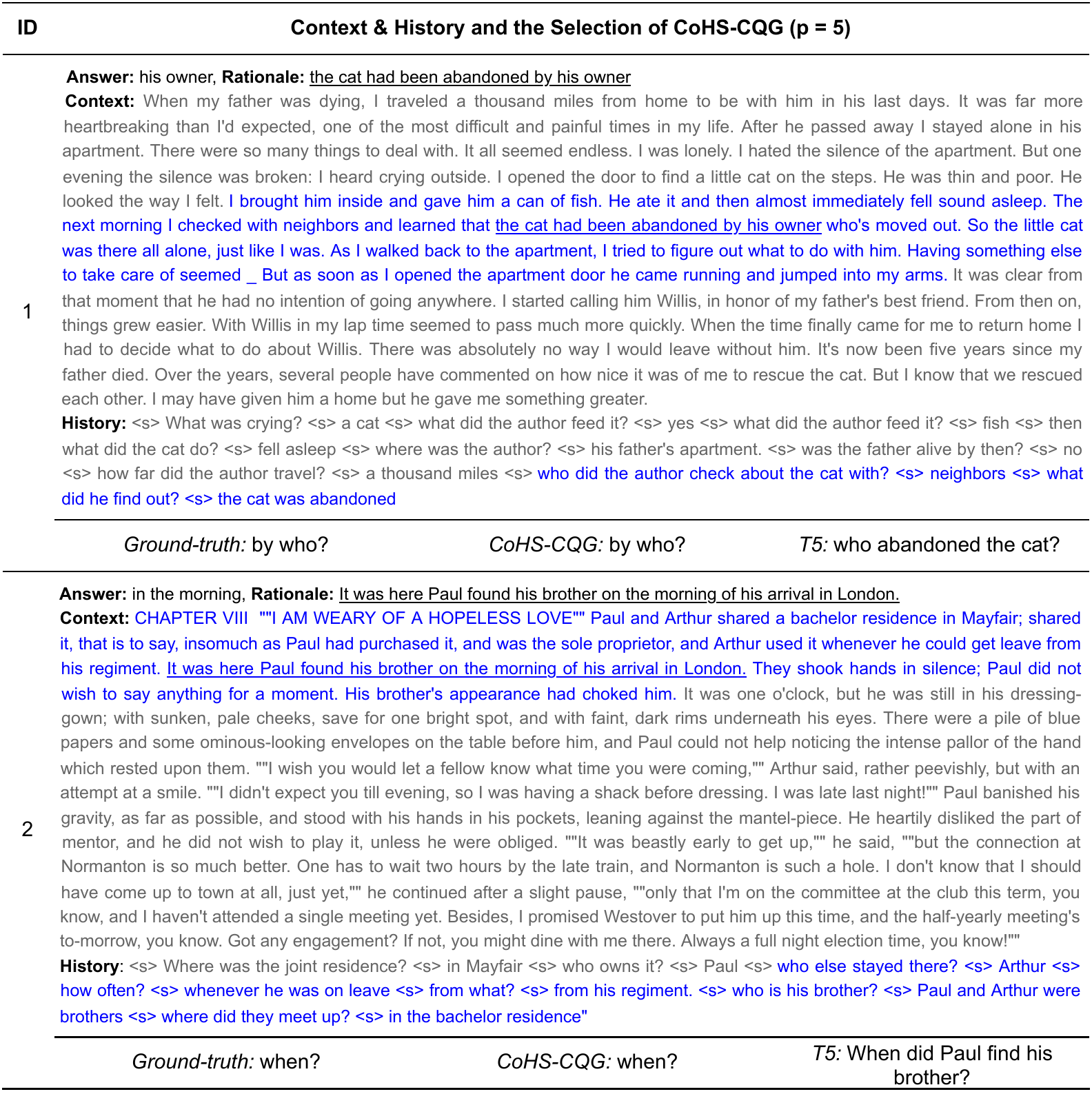}
    \vspace{-1.5cm}
     \caption{Case studies on the CoQA validation set and the results of T5 (with full context and history) and CoHS-CQG (with shorten context and history). The texts of rationales are underlined. The selected texts in context and history by our \emph{CoHS (p = 5)} module are highlighted in blue.}
    \label{appendix:case-studies-examples}
\end{figure*} 

In \Cref{tab:ablation-previous-turn}, we study \change{with the full} context, how the number of previous history turns affect the performance of the model on our validation set. We can observe that with the full context, the settings of previous history turns $k=2$ and $k=3$ achieve the best performances on ROUGE-L and BLEU-4, respectively. Compared to the performances in \Cref{tab:ablation-dynamic-history}, it indicates that \change{dynamically selecting} instead of fixing the number of previous turns indeed improves the performance.

\begin{table}[t!]
\centering
\resizebox{0.45\textwidth}{!}{%
\begin{tabular}{lccc}
    \hline
    Model & Flu. & C-Align & Ans. \\
    \midrule \multicolumn{4}{c}{\textbf{Answer-unaware}} \\
    \emph{CoHS-CQG w/o AE} & 2.13 & 1.76 & 1.64 \\
    \emph{CoHS-CQG w/o QF} & 2.16 & 1.98 & 2.02 \\
    \hline
    \emph{CoHS-CQG} (\emph{Ours}) & \textbf{2.74} & \textbf{2.58} & \textbf{2.60} \\
    \hline
    Krippendorff's $\alpha$ & 0.82 & 0.79 & 0.77 \\
    \hline   \end{tabular}}

\caption{Human evaluation \change{results} for the ablation studies of AE and QF modules on the validation set of CoQA. ``Krippendorff's $\alpha$'' shows the inter-annotator agreement. \textit{Flu.}: Fluency, \textit{C-Align}: Conversational Alignment, \textit{Ans.}: Answerability. }

\label{tab:human-evaluation-ablation-study}
\vspace{-3mm}
\end{table}

\subsection{Ablation Studies} \label{sec:ablation-study}

\paragraph{Ablation of Answer-span Extractor (AE)} \label{sec:ablation-answer-span-extractor}
\change{We conduct an ablation study for the \emph{Answer-span Extractor (AE)} (\emph{CoHS-CQG w/o AE})\change{,} in which we replace the predicted answer span $a_n$ with the rationale $r_n$ (a sentence in the context $C$). The results are shown in \Cref{tab:human-evaluation-ablation-study}. Note that in this experiment, we also remove the \emph{Question Filtering (QF)}. As expected, the \emph{Answerability} and \emph{Conversational Alignment} drop significantly, which is explainable since the rationale $r_n$ may contain redundant information, thus it is not suitable to be $r_n$.}

\paragraph{Ablation of Question Filtering (QF)} \label{sec:ablation-question-filtering}
\change{We study the ablation of the Question Filtering (\emph{CoHS-CQG w/o QF})\change{,} in which we use all the generated questions. The results are shown in \Cref{tab:human-evaluation-ablation-study}. As we can see, the \emph{Answerability} and \emph{Conversational Alignment} decrease significantly. We observe that without QF, there may have \change{been} turns in which the questions are similar with the same answers, which further proves the necessity of this module.}

\subsection{Case study: Effectiveness of \emph{CoHS}} \label{appendix:why-low-performance}

When carefully studying the \change{performances} of T5 and BART, we observe that the key for these models to gain high scores on n-gram automatic metrics\change{, such as BLEU and ROUGE,} is \emph{focusing on the history to optimise the conversational alignment}. We argue that, with a long context and the whole conversation history, the input \change{likely tends to} distract the attention of the models on the \change{given conversation} history. The models in these cases mostly focus on the given answer and rationale to generate the question, rather than \change{highly} focusing on the history. \Cref{appendix:case-studies-examples} lists some of the examples whereby we draw the above conclusion. 
% The examples of the above conclusions are in \Cref{appendix:case-studies-examples}. 

Considering the first example in \Cref{appendix:case-studies-examples}, we observe that with the full context and history, the T5 model mostly relies on the rationale ``\emph{the cat had been abandoned by his owner}'' to generate the question ``\emph{who abandoned the cat?}''. Although the question is somehow aligned with the given conversation history, it is not close enough to the gold question ``\emph{by who?}''. The other two examples in \Cref{appendix:case-studies-examples} are also the same, and we observe a lot of cases that are similar to them. 
We argue that in order to generate such questions like ``\emph{by who?}'', intuitively, the model should pay significant attention to the conversation history to optimise the conversational alignment. By \change{inputting to} the model the shortened context and history, we can see that the generated questions in \Cref{appendix:case-studies-examples} by \emph{CoHS-CQG} indeed change, and they are exactly the same as the \change{ground-truth} questions. This improvement reflects that training the model such as T5 with the shortened context and history samples \change{indeed guides} the model to optimising \change{more on the conversational alignment property} instead of just heavily focusing on the target answer and rationale. 

\section{Conclusion}

This paper presents \emph{CoHS-CQG}, a two-stage framework for CQG, which adopts a \emph{CoHS} module to dynamically select relevant context and conversation history for generating the question in the current turn. 
% For reproducibility, the \emph{CoHS} module can be plugged into any CQG model easily. 
Experimental results on CoQA demonstrate that the proposed \emph{CoHS-CQG} 
%\bwc{I suggest removing ``model'' here since commonly we propose CoHS-CQG as a framework in this paper.} 
achieves state-of-the-art performances in both answer-aware and answer-unaware settings. Our extensive analysis and studies show the effectiveness of \emph{CoHS} in improving the CQG models. In future work, we will focus on how to select the contiguous question-worthy content from the paragraph by reasoning.

% We present \emph{CoHS-CQG}, a two-stage CQG model, which adopts a novel module, \emph{CoHS}, to shorten the context and the history before inputting them to the QG model. The proposed module, \emph{CoHS}, can be plugged into any model, which makes it easily reproducible. We establish new strong state-of-the-art performances on answer-aware and answer-unaware CQG. By our thorough study and analysis, we show the selection capacity of \emph{CoHS} and thus its effectiveness on helping the models to optimise the generated questions.

\section*{Acknowledgements}
\change{This research has been supported by the Institute for Infocomm Research of A*STAR (CR-2021-001).} \change{We would like to thank Tran Anh Tai for his support, and anonymous reviewers for their valuable feedback on how to improve the paper.}

\bibliography{anthology,custom}
\bibliographystyle{acl_natbib}

\clearpage

\appendix
\section{Appendix} \label{sec:appendix}

\subsection{Comparison with Static Context Selection} \label{appendix:select-five-sentences}

% When $H_n = \emptyset$, then $H_{sub} = \emptyset$, and $C_{sub}$ is \emph{five} sentences around $c_s$, see \Cref{appendix:select-five-sentences}.
To compare our dynamic context selection strategy with a static way, we simply select \emph{five} context sentences around $c_s$ in the context $C = \{c_1, c_2, ..., c_m\}$ since \Cref{tab:ablation-dynamic-context} shows that around more than four sentences on average achieves the best performance. To this end, we consider a simple heuristic method as below. If $3 \leq s \leq m - 2$, $C_{sub} = \{c_{s-2}, c_{s-1}, c_s, c_{s+1}, c_{s+2}\}$; else if $s \leq 2$, $C_{sub} = \{c_1, c_2, c_3, c_4, c_5\}$; and if $s \geq m - 1$, $C_{sub} = \{c_{m-4}, c_{m-3}, c_{m-2}, c_{m-1}, c_{m}\}$. 
We then select five sentences in the context by this way and use the whole conversation history. The result shows that the T5$_{base}$ model yields 17.24 of BLEU-4 and 47.02 of ROUGE-L, which slightly outperforms the T5$_{base}$ baseline using the full context in \Cref{tab:main-experiment}.

\subsection{Human Rating System} \label{ssec:human-rating-system}
\begin{figure}[t!]
\centering

\includegraphics[width=8cm]{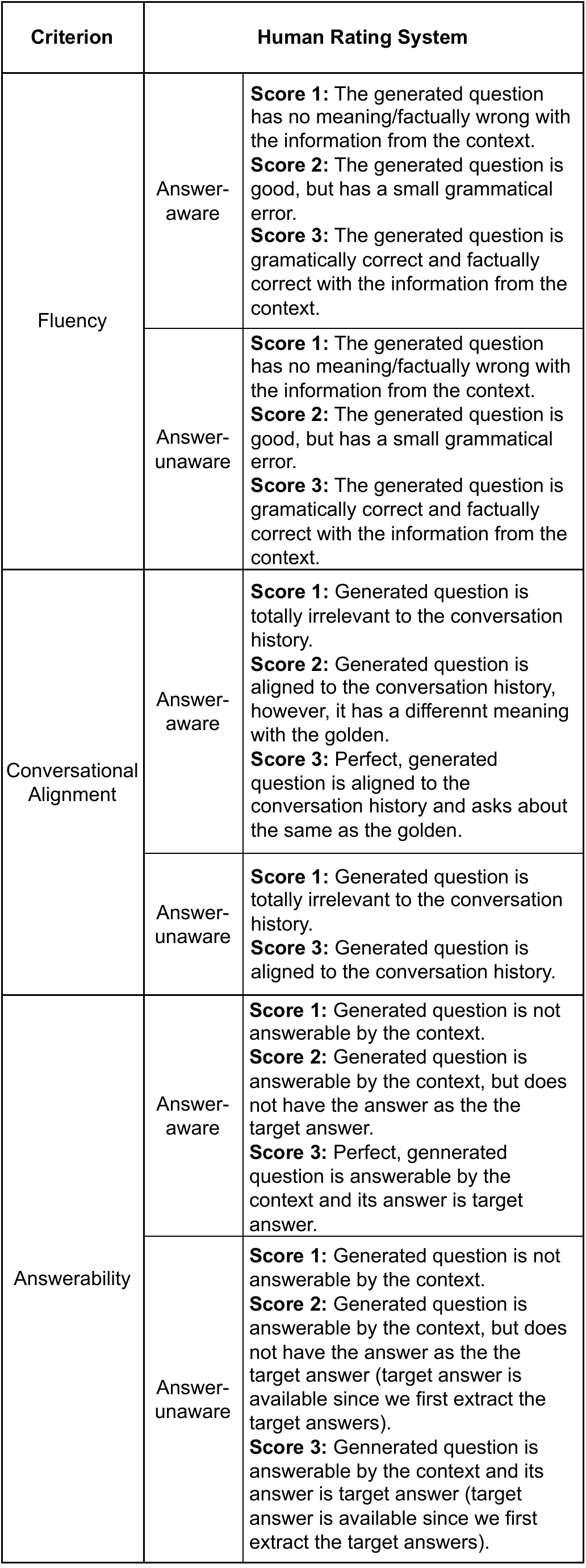}
\vspace{1mm}
{\begin{tabular}{p{15cm}}
\end{tabular}}
\vspace{-3mm}
\caption{\small Human Rating System
}
\vspace{-3mm}
\label{fig:human-rating-system}
\end{figure}

In this section, we describe how our annotators are instructed to give the points in three criteria \emph{Fluency}, \emph{Conversational Alignment}, and \emph{Answerability}. There are three main notes. First, \emph{Fluency} measures not only the grammatical correctness, but also measures the meaning and factual correctness of the question with the given context. Second, in the answer-unaware setting, as there is no golden history, we do not define the \emph{Score 2} in the \emph{Conversational Alignment} as in the answer-aware setting. Third, for the \emph{Answerability} criterion in the answer-unaware setting, the target answer and target rationale are unavailable. However, since our approach first selects the rationale, and then extracts the candidate answers from it to generate the questions, we still evaluate the quality of our questions (\emph{Score 2, 3}) with the selected candidate answers by the \emph{Question Filtering} module (\Cref{conqg-framework}). For the details, see \Cref{fig:human-rating-system}.

% \vspace{-3mm}
\end{document}